\documentclass[11pt]{article}
\usepackage{arxiv}
\usepackage[numbers]{natbib}
\usepackage{amsmath,amssymb}
\usepackage{booktabs}
\usepackage{tabularx}
\usepackage{makecell}
\usepackage{tabularx}
\usepackage{makecell}
\usepackage{graphicx}
\usepackage{silence}
\title{Sonny: Breaking the Compute Wall in Medium-Range Weather Forecasting}

\author{
  Minjong Cheon\\
  Department of Computer Science and Engineering, Sejong University\\
  Seoul, South Korea\\
  \texttt{jmj2316@sejong.ac.kr}
}

\begin{document}
\maketitle
\begin{abstract}
Weather forecasting is a fundamental problem for protecting lives and infrastructure from high-impact atmospheric events. Recently, data-driven weather forecasting methods based on deep learning have demonstrated strong performance, often reaching accuracy levels competitive with operational numerical systems. However, many existing models rely on large-scale training regimes and compute-intensive architectures, which raises the practical barrier for academic groups with limited compute resources. Here we introduce Sonny, an efficient hierarchical transformer that achieves competitive medium-range forecasting performance while remaining feasible within reasonable compute budgets. At the core of Sonny is a two-stage StepsNet design: a narrow slow path first models large-scale atmospheric dynamics, and a subsequent full-width fast path integrates thermodynamic interactions. To stabilize medium-range rollout without an additional fine-tuning stage, we apply EMA during training. On WeatherBench2, Sonny yields robust medium-range forecast skill, remains competitive with operational baselines, and demonstrates clear advantages over FastNet, particularly at extended tropical lead times. In practice, Sonny can be trained to convergence on a single NVIDIA A40 GPU in approximately 5.5 days.
\end{abstract}

\section{Introduction}

Weather forecasting is essential for protecting lives, infrastructure, and economic systems from high-impact atmospheric events \cite{IPCC2021,ahn2023searching,bodnar2025foundation}. As climate variability intensifies, improving both forecast skill and accessibility has become a central scientific and societal priority \cite{IPCC2021}. Recent advances in data-driven Numerical Weather Prediction (NWP) show that CNN-based models, Graph Neural Networks (GNNs), and Transformers can match, and in some settings exceed, the skill of traditional physics-based systems \cite{cheon2025modernizing,cheon2024karina,Bi2023Pangu,Pathak2022FourCastNet,Lam2023GraphCast}. However, many Large Weather Models (LWMs) still depend on computationally intensive training pipelines that require large-scale TPU/GPU infrastructure \cite{Bi2023Pangu,Lam2023GraphCast,dunstan2025fastnet}. This training-time compute barrier can substantially limit broader academic participation in developing next-generation global forecasting models \cite{khadir2025democracy}.

To address this accessibility gap, we target a compact model that can be trained with limited compute while still delivering reliable medium-range global forecasts. Existing studies have primarily emphasized either maximum skill at large scale or isolated efficiency gains on specific setups \cite{Lam2023GraphCast, bonev2023sfno, han2024fengwu, bonev2025fourcastnet}, leaving a gap for a compute-accessible baseline that jointly prioritizes forecast quality, training stability, and reproducibility in a single end-to-end framework.

To fill this gap, we present Sonny, an efficient Small Weather Models (SWM) based on the StepsNet architecture \cite{han2025step}. Sonny is designed to achieve competitive medium-range forecast skill at practical training cost, and it is developed as a reproducible baseline for broader academic use. Recent efficient SWM efforts, including ArchesWeather \cite{couairon2024archesweather} and FastNet \cite{dunstan2025fastnet}, also highlight the importance of balancing forecast quality with training efficiency under constrained compute.

Our key contributions are threefold. First, we introduce Sonny, an efficient SWM for medium-range global weather forecasting under limited compute budgets. Second, we provide a practical training recipe aimed at stable medium-range rollout without expensive large-scale infrastructure. Third, we demonstrate on WeatherBench2 that Sonny achieves competitive performance against HRES and AI baselines while remaining trainable on a single NVIDIA A40 GPU in approximately 5.5 days.

\section{Method}

\subsection{Data}

We trained and evaluated Sonny using the WeatherBench2 (WB2) dataset \cite{rasp2024weatherbench2} with 6-hourly temporal resolution and \(1.5^{\circ}\) spatial resolution (\(121 \times 240\)). Our input includes four surface variables and five atmospheric variables defined on 13 pressure levels (50, 100, 150, 200, 250, 300, 400, 500, 600, 700, 850, 925, and 1000 hPa). We split the data chronologically into 1979--2018 for training, 2019 for validation, and 2022 for testing. The full variable list is summarized in Table~\ref{tab:data_variables}.

\begin{table}[htbp]
    \caption{List of ERA5 atmospheric variables used in Sonny, corresponding short names, vertical pressure levels in hectopascals (hPa), and standard units of measurement.}
    \centering
    \small
    \begin{tabularx}{\columnwidth}{>{\raggedright\arraybackslash}X l >{\raggedright\arraybackslash}X l}
        \toprule
        \textbf{Variable name} & \textbf{Short name} & \textbf{Vertical levels (hPa)} & \textbf{Units} \\
        \midrule
        Zonal wind & U & \makecell[l]{1000, 925, 850, 700, 600, 500, 400,\\300, 250, 200, 150, 100, 50} & m/s \\
        \addlinespace
        Meridional wind & V & \makecell[l]{1000, 925, 850, 700, 600, 500, 400,\\300, 250, 200, 150, 100, 50} & m/s \\
        \addlinespace
        Temperature & T & \makecell[l]{1000, 925, 850, 700, 600, 500, 400,\\300, 250, 200, 150, 100, 50} & K \\
        \addlinespace
        Specific humidity & Q & \makecell[l]{1000, 925, 850, 700, 600, 500, 400,\\300, 250, 200, 150, 100, 50} & kg/kg \\
        \addlinespace
        Geopotential & Z & \makecell[l]{1000, 925, 850, 700, 600, 500, 400,\\300, 250, 200, 150, 100, 50} & m\(^2\)/s\(^2\) \\
        \addlinespace
        2m temperature & T2m & \multicolumn{1}{c}{-} & K \\
        Mean sea level pressure & MSLP & \multicolumn{1}{c}{-} & Pa \\
        Surface air pressure & SP & \multicolumn{1}{c}{-} & Pa \\
        \makecell[l]{10m zonal wind} & U10 & \multicolumn{1}{c}{-} & m/s \\
        \makecell[l]{10m meridional wind} & V10 & \multicolumn{1}{c}{-} & m/s \\
        \bottomrule
    \end{tabularx}
    \normalsize
    \label{tab:data_variables}
\end{table}

\subsection{Randomized Dynamics Forecasting}

Sonny was trained to predict the atmospheric dynamics \(\Delta_{\delta t} = X_{\delta t} - X_0\) rather than absolute states, representing the evolution of weather conditions over a time interval \(\delta t\). Unlike conventional models that rely on a fixed interval (e.g., 6 hours), we used a randomized dynamics forecasting objective. During training, \(\delta t\) was sampled from a discrete uniform distribution \(P(\delta t) \sim \mathcal{U}\{6, 12, 24\}\) hours \cite{nguyen2024scaling}.

\subsection{Pressure-Weighted Loss Function}

To prioritize variables with the highest societal impact, we employed a physics-based pressure-weighted loss. This function uses atmospheric pressure as a proxy for density, assigning higher weights to variables near the Earth's surface \cite{nguyen2024scaling}. The final loss function incorporated both this pressure weighting \(w(v)\) and a latitude-weighting factor \(L(i)\) to account for the non-uniformity of the spherical grid:

\[
\mathcal{L}(\theta) = \mathbb{E} \left[ \frac{1}{VHW} \sum_{v=1}^{V} \sum_{i=1}^{H} \sum_{j=1}^{W} w(v)L(i) \left( \Delta \hat{v}_{ij}^{\delta t} - \Delta v_{ij}^{\delta t} \right)^2 \right]
\]

\subsection{Efficient Training via EMA}

One of bottlenecks in state-of-the-art weather models is the extensive fine-tuning required to stabilize long-term iterative forecasts. In Sonny, we bypassed this computationally expensive fine-tuning phase by implementing an Exponential Moving Average (EMA) during the primary training stage \cite{polyak1992acceleration,tarvainen2017meanteacher,grill2020byol}. We applied a decay factor of 0.9 to the model weights, which effectively smooths the optimization landscape and enhances the model's robustness. This approach served as a highly efficient alternative to traditional fine-tuning, allowing the model to achieve superior stability in iterative rollouts without additional temporal overhead. In our setup, the full training process converged in approximately 5.5 days on a single NVIDIA A40 GPU, with peak GPU memory usage of about 32 GB.

\subsection{Sonny}

Sonny is a hierarchical weather forecasting model based on the StepsNet architecture, designed to exploit the distinct physical characteristics of atmospheric variables. The model's core is the Variable-Aware Embedding module, which decouples input variables into two physically informed categories: a Dynamics group (\(U, V, Z, P\)) and a Thermodynamics group (\(T, Q\)). This separation is motivated by the physical hierarchy of the atmosphere, where kinematic drivers (dynamics) dictate the large-scale evolution that thermodynamic states subsequently follow.

This variable-aware separation facilitates a dual-phase processing pipeline. In Step 1 (Slow Path), the Dynamics group is embedded into a \(d_1\)-dimensional space and processed through a specialized sequence of \(N_1\) Transformer blocks. By restricting the hidden dimension to \(d_1\), Sonny can afford a deeper network architecture to distill the fundamental atmospheric backbone without the quadratic memory overhead typically associated with processing all variables simultaneously. This stage focuses on capturing long-range spatial dependencies and the structural evolution of the pressure fields.

Subsequently, these refined dynamic features (\(y_1\)) are concatenated with the raw Thermodynamics embeddings (\(d_2\)) to form a comprehensive representation. In Step 2 (Fast Path), the integrated features are processed through \(N_2\) full-width blocks of dimension \(d = d_1 + d_2\). This stage is designed to capture complex, non-linear interactions---such as moisture advection and latent heat release---where the pre-processed dynamic backbone serves as a structural guide for the thermodynamic variables.

To handle various lead times, Sonny incorporates a Randomized Dynamics Conditioning mechanism. The time interval \(\delta t\) is embedded and injected into each Transformer block via adaptive Layer Normalization (adaLN-Zero) \cite{peebles2023dit,han2025step}. Specifically, for Step 1, the time embedding is linearly projected to \(d_1\), while Step 2 utilizes the full-scale embedding. This hierarchical approach significantly optimizes computational efficiency; by deferring the processing of thermodynamics to Step 2, Sonny reduces the total FLOPs while maintaining high fidelity in capturing multi-scale atmospheric dependencies.

\begin{figure}[htbp]
\centering
\includegraphics[width=0.95\columnwidth]{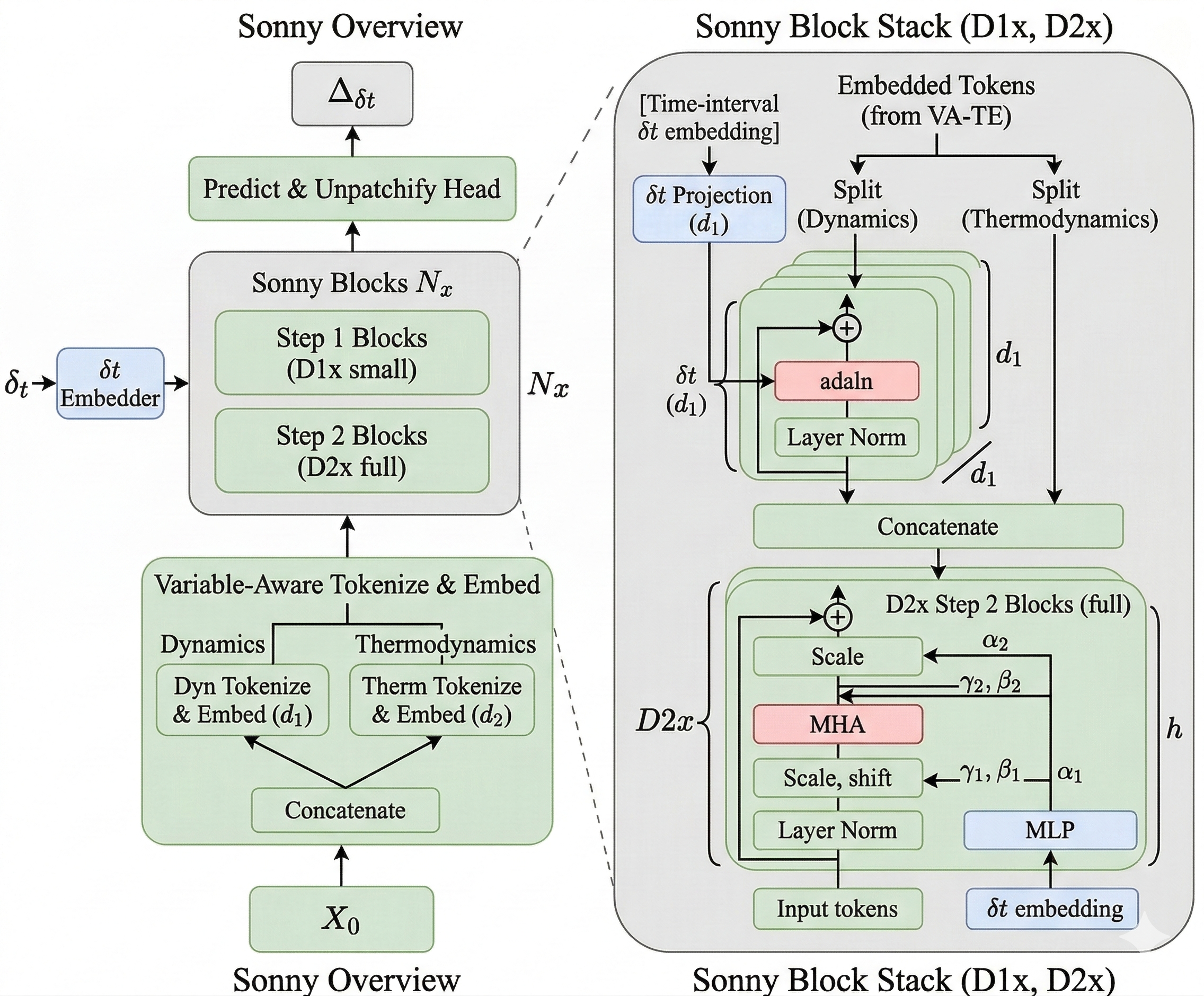}
\caption{Overview of Sonny (StepsNet) with variable-aware embedding. Step 1 models dynamics, and Step 2 fuses thermodynamics for efficient, accurate forecasting.}
\label{fig:sonny_architecture}
\end{figure}

Finally, the output of the Step 2 blocks is passed through a Final Layer and an Unpatchify operation to reconstruct the predicted atmospheric state \(\Delta_{\delta t}\) at the target interval. The overall two-stage pipeline is illustrated in Figure~\ref{fig:sonny_architecture}. Detailed training hyperparameters are summarized in Table~\ref{tab:implementation_details}.

\begin{table}[htbp]
\caption{Training configuration for Sonny.}
\label{tab:implementation_details}
\centering
\begin{tabular}{ll}
\toprule
Item & Value \\
\midrule
Backbone setting & ViT-S configuration \\
Number of parameters & 20.5M \\
\multicolumn{2}{l}{\textbf{Optimizer}} \\
Class & AdamW \\
Learning rate & 5e-4 \\
\(\beta_1\) & 0.9 \\
\(\beta_2\) & 0.95 \\
Weight decay & 1e-5 \\
Batch size & 16 \\
Total epochs & 50 \\
\bottomrule
\end{tabular}
\end{table}

\begin{table*}[htbp]
\caption{Compute and deployment comparison across representative weather models. Entries marked with $^*$ are reported from source papers under their own settings.}
\label{tab:compute_comparison}
\centering
\begin{tabular}{lccccc}
\toprule
Model & Params (M) & MACs (G) & Training Hardware & Training Time & Latency (ms) \\
\midrule
\multicolumn{6}{l}{\textbf{721 $\times$ 1440 (0.25$^{\circ}$)}} \\
Fengwu & 153.49 & 132.83 & 32 $\times$ A100 & 17 days & 227.0 \\
FourCastNet & 79.60 & 111.62 & 64 $\times$ A100 & 16 hours & 164.0 \\
GraphCast & 36.7 & 1639.26 & 32 $\times$ TPUv4 & 4 weeks & - \\
Pangu-Weather & 256 & 142.39 & 192 $\times$ V100 & 64 days & 275.0 \\
\midrule
\multicolumn{6}{l}{\textbf{128 $\times$ 256 ($\sim$1.4$^{\circ}$)}} \\
VA-MoE & 665.37 & - & 16 $\times$ A100 & 12 days$^*$ & 269.0 \\
OneForecast & 24.76 & 509.27 & 16 $\times$ A100 & 8 days$^*$ & 663.0 \\
STCast & 654.82 & 436.12 & 16 $\times$ A100 & 5 days$^*$ & 115.0 \\
EMFormer & 157.90 & 102.54 & 16 $\times$ A100 & 60 hours$^*$ & 98.3 \\
\midrule
\multicolumn{6}{l}{\textbf{121 $\times$ 240 (1.5$^{\circ}$)}} \\
Sonny & 20.50 & 96.81 & 1 $\times$ A40 & 5.5 days & 125.48 \\
\bottomrule
\end{tabular}
\end{table*}

Table~\ref{tab:compute_comparison} summarizes model-size and efficiency metrics across representative baselines and our 121 $\times$ 240 (1.5$^{\circ}$) setting \cite{chen2026emformer,cheon2024karina}.

\section{Experimental Results and Analysis}

\subsection{Impact of EMA on Weather Forecasting}

\begin{figure}[htbp]
\centering
\includegraphics[width=0.98\columnwidth]{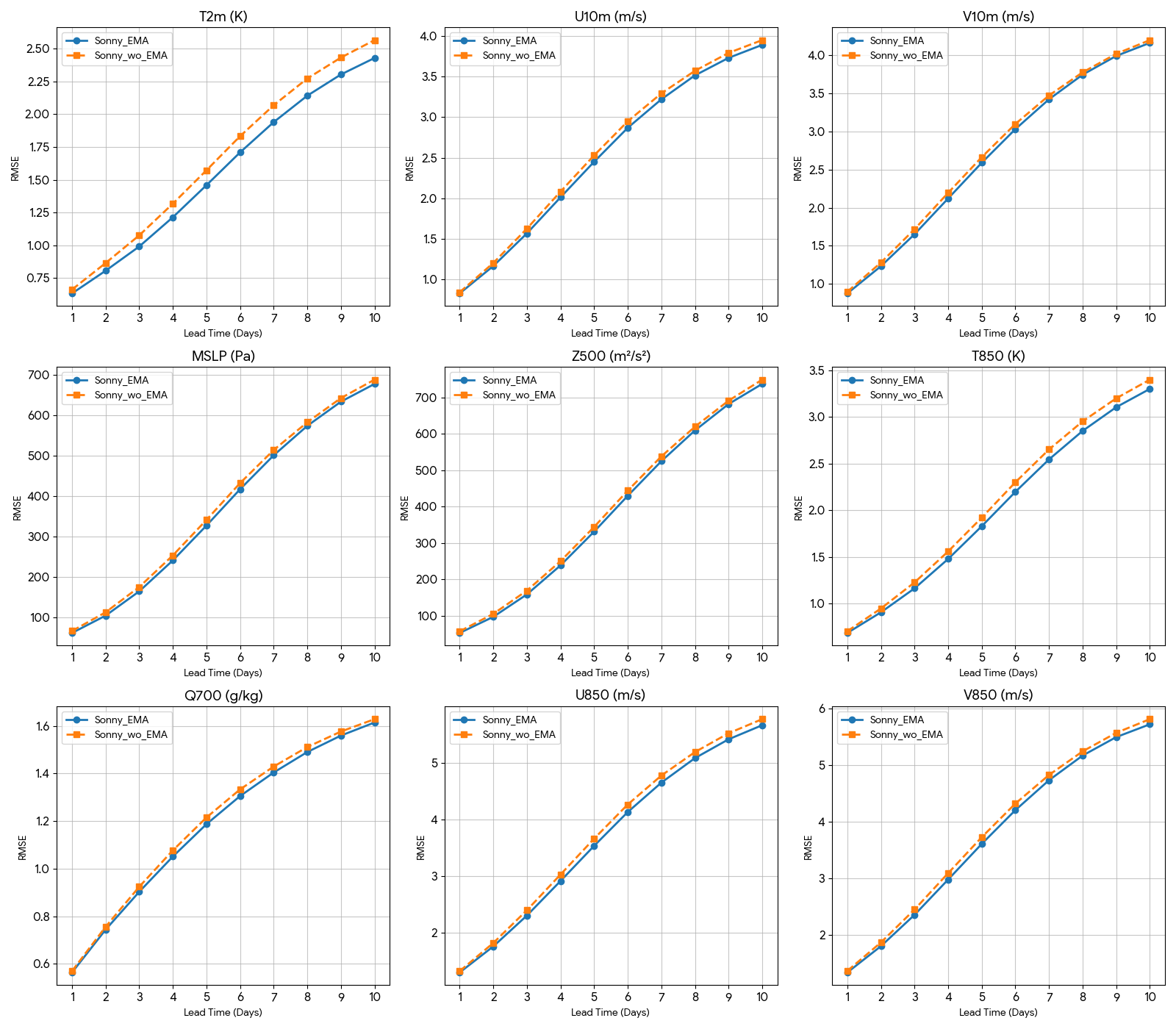}
\caption{RMSE over lead time for nine variables, comparing Sonny with EMA (Sonny\_EMA) and without EMA (Sonny\_wo\_EMA).}
\label{fig:effect_ema}
\end{figure}

Figure~\ref{fig:effect_ema} demonstrates a comprehensive quantitative evaluation of the Sonny model's predictive performance, measured by Root Mean Square Error (RMSE) over a 10-day lead time. To enhance training stability and model generalization, EMA with a decay rate of 0.9 was applied to the model weights during the training phase. As illustrated across the subplots, the model variant utilizing EMA (Sonny\_EMA) consistently achieves lower RMSE values compared to the baseline model (Sonny\_wo\_EMA) across all nine meteorological variables. Specifically, applying EMA yields an overall average error reduction of approximately 3.34\% across the entire forecast period. This performance gain is particularly notable in the short-to-medium range, peaking at nearly 4.75\% error reduction on Day 3, and maintains a solid \textasciitilde2.00\% improvement even toward Day 10. The results clearly indicate that maintaining an exponential moving average of the weights (with \(\alpha = 0.9\)) effectively mitigates error growth and significantly improves the forecasting accuracy of the model.

\subsection{Comparative Performance Evaluation}

\begin{figure}[htbp]
\centering
\includegraphics[width=0.98\columnwidth]{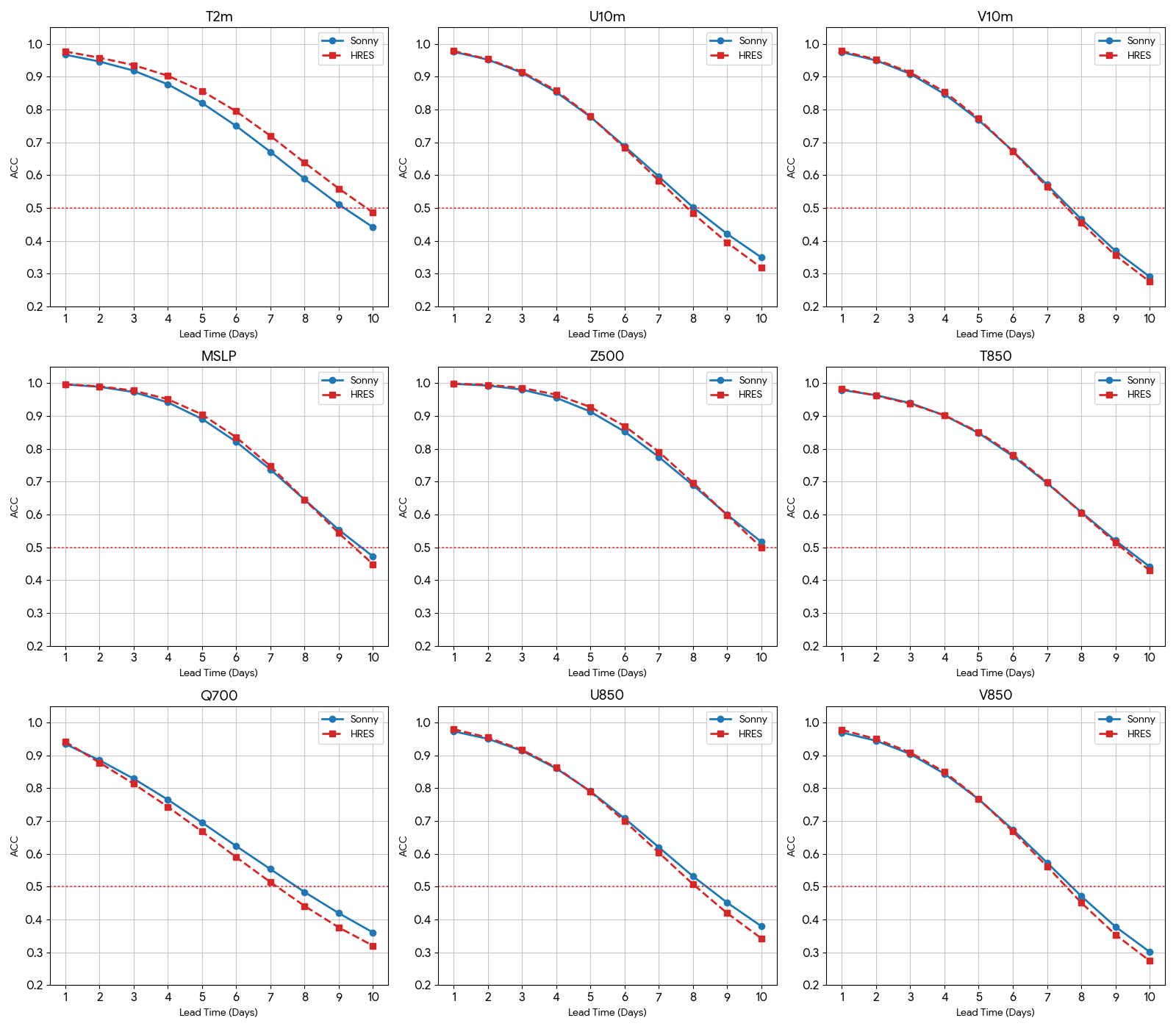}
\caption{ACC over a 10-day lead time for nine variables, comparing Sonny and HRES. The dotted line marks ACC = 0.5.}
\label{fig:comparison_hres}
\end{figure}

Figure~\ref{fig:comparison_hres} compares Sonny directly against HRES over a 10-day horizon. Sonny remains competitive across key variables while offering a substantially more efficient inference pathway.

\begin{figure}[htbp]
\centering
\includegraphics[width=0.98\columnwidth]{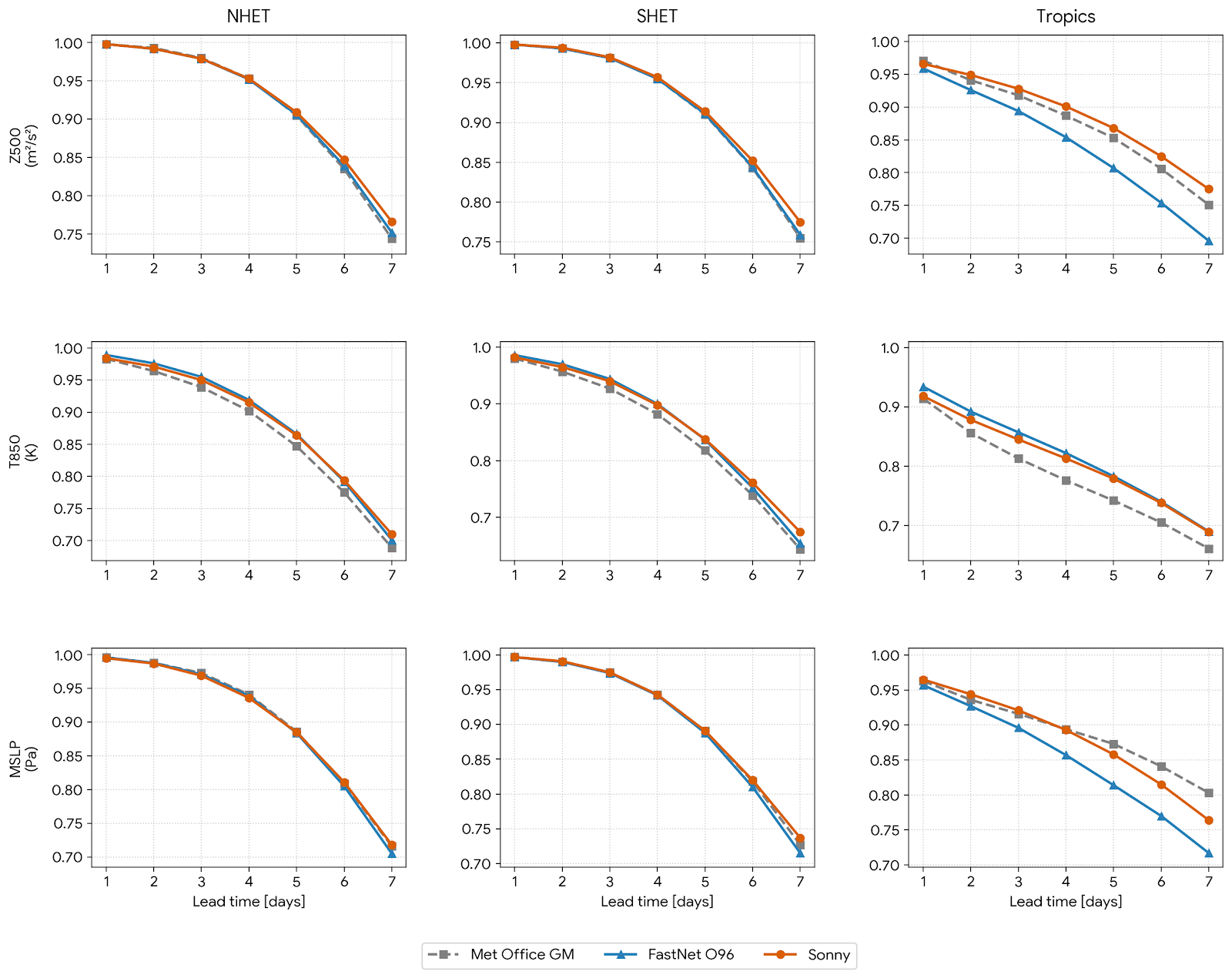}
\caption{ACC for Z500, T850, and MSLP up to day 7, comparing Sonny, Met Office GM, and FastNet O96 over NHET, SHET, and the Tropics.}
\label{fig:comparison_multi_baselines}
\end{figure}

Figure~\ref{fig:comparison_multi_baselines} extends the comparison to Met Office GM and FastNet O96 across NHET, SHET, and the Tropics. In the extratropical regions (NHET and SHET), the Sonny model demonstrated robust baseline performance. For Z500 and MSLP, Sonny maintained high ACC scores that are strictly comparable to both Met Office GM and FastNet O96 across all lead times. Notably, in T850 forecasts, Sonny exhibited a lower rate of performance degradation compared to FastNet O96 as lead time increased, closely matching the accuracy of the physics-based Met Office GM.

The most significant performance improvement was observed in the extended lead times within the Tropics. While other models experienced rapid degradation in prediction skill due to the complex atmospheric dynamics of the region, the Sonny model achieved a substantial increase in ACC across all three variables compared to FastNet O96. Specifically, for Z500 and T850 beyond day 4, Sonny significantly widened the performance gap with FastNet O96, maintaining high predictive skill that is either comparable or superior to the Met Office GM. Overall, Sonny maintains strong competitiveness across baselines while preserving a substantially more efficient inference pathway.

\subsection{Case Study: 2022 Typhoon Nanmadol}

\begin{figure*}[htbp]
\centering
\includegraphics[width=0.98\textwidth]{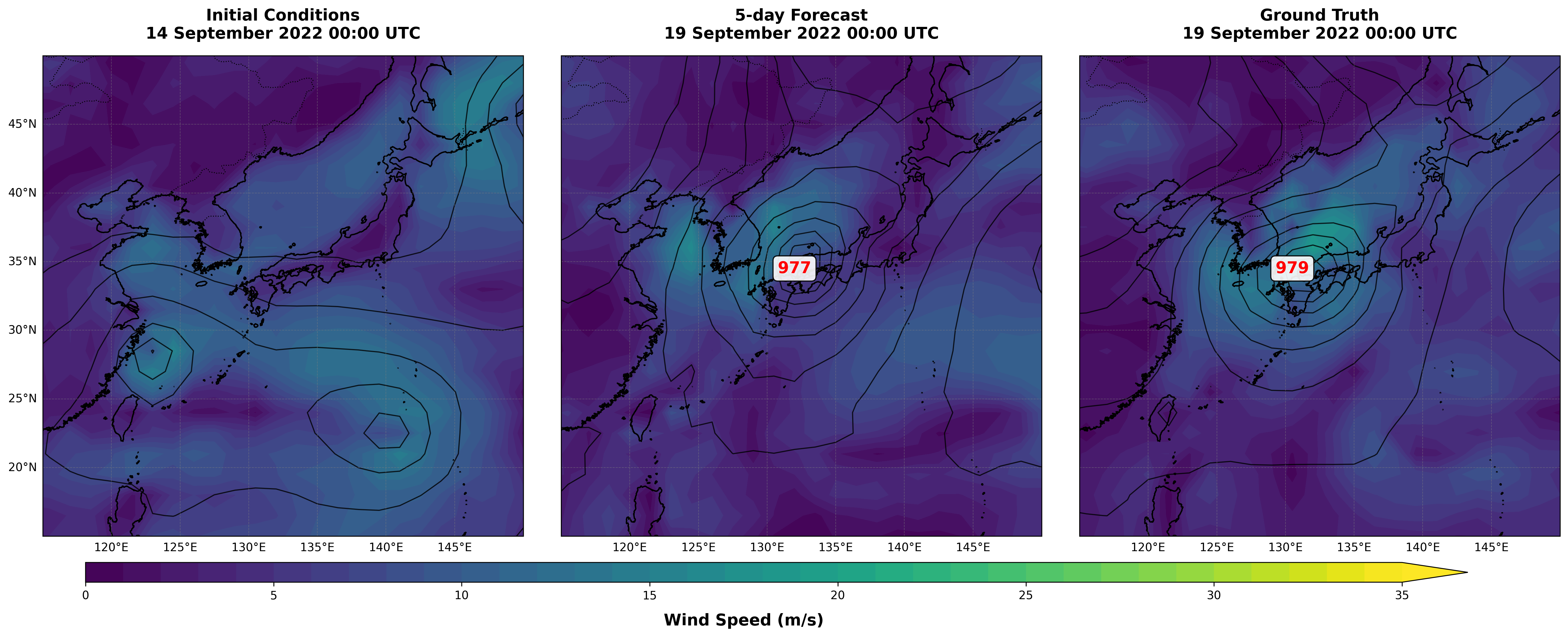}
\caption{Typhoon Nanmadol case (September 2022): (a) initial state, (b) 120-hour Sonny forecast, and (c) ERA5 reference. Shading shows 10 m wind speed, and contours show MSLP.}
\label{fig:nanmadol_regional}
\end{figure*}

Figure~\ref{fig:nanmadol_regional} presents regional forecast fields for Typhoon Nanmadol (September 2022), used here as a representative high-impact event for medium-range evaluation. Forecasting was initialized at 00:00 UTC on September 14, 2022, and integrated for 120 hours (5 days) with 6-hour output intervals. The model performs strongly in both track and intensity prediction. At 120-hour lead time, the track error is about 137.5 km, comparable to reported results from leading AI weather models, including Pangu-Weather and GraphCast, at similar forecast horizons. For intensity, the predicted minimum central pressure is 977 hPa, closely aligned with the ERA5 reference value of 979 hPa (error: approximately 1.4 hPa). These results suggest that the proposed architecture mitigates the over-smoothing tendency commonly observed in deep-learning weather forecasting systems.

\subsection{Case Study: 2022 Winter Storm Elliott}

\begin{figure*}[htbp]
\centering
\includegraphics[width=0.98\textwidth]{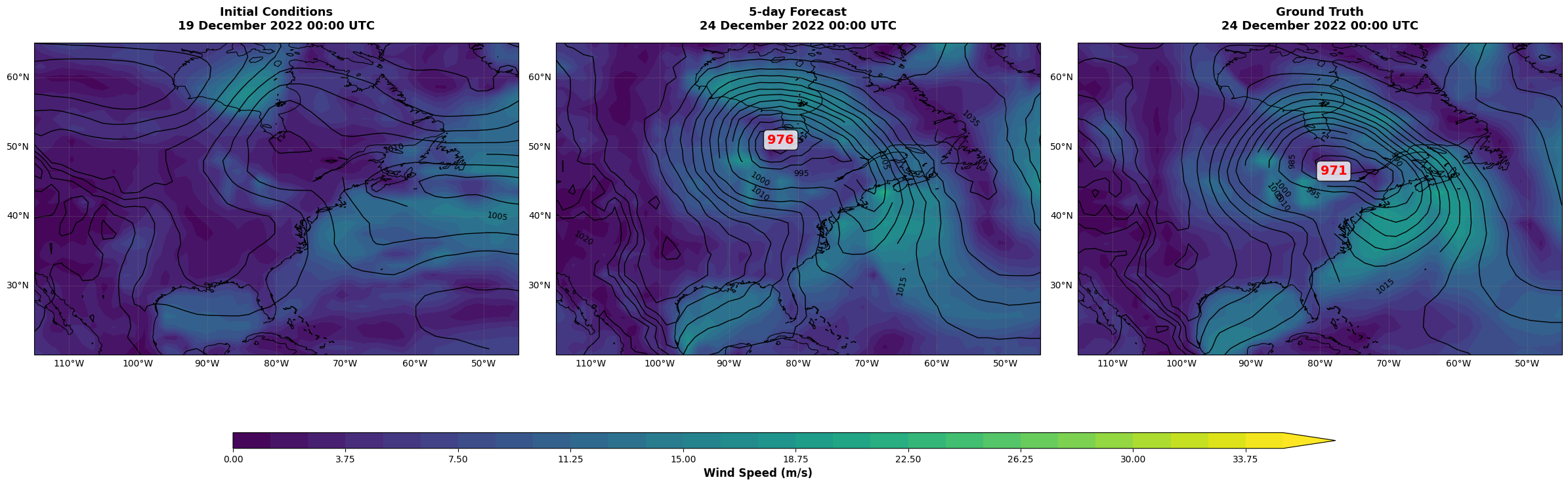}
\caption{Winter Storm Elliott case (December 2022): (a) initial state, (b) 120-hour Sonny forecast, and (c) ERA5 reference. Shading shows 10 m wind speed, and contours show MSLP.}
\label{fig:elliott_regional}
\end{figure*}

Figure~\ref{fig:elliott_regional} presents regional forecast fields for Winter Storm Elliott (December 2022), selected as a representative extreme extratropical event. The forecast was initialized at 00:00 UTC on December 19, 2022, and rolled out for 120 hours (5 days) at 6-hour intervals. Despite the coarse \(1.5^{\circ}\) spatial resolution, the model captures this rapid intensification well. At 120-hour lead time, the predicted minimum central pressure is 976.7 hPa, close to the ERA5 reference value of 971.8 hPa. This 4.9 hPa intensity error indicates that Sonny captures deep synoptic-scale structure while mitigating over-smoothing behavior commonly observed in lower-resolution deep-learning weather models. In addition, the model locates the storm center with a spatial error of \(4.5^{\circ}\) in both latitude and longitude, corresponding to a shift of three grid cells. Given the 5-day horizon and strongly nonlinear evolution of this bomb cyclone, these results highlight Sonny's ability to model complex atmospheric dynamics and provide reliable early-warning signals for high-impact winter storms.
\section{Conclusion and Future Work}

This work proposes Sonny, an efficient deep learning model for medium-range weather forecasting. We show that a hierarchical transformer with a physically informed variable split can achieve competitive performance. Our core design, the two-stage StepsNet pipeline with randomized dynamics forecasting, trains the model across multiple time intervals and enables flexible lead-time inference. Experimental results on WeatherBench2 demonstrate strong medium-range skill and competitive performance against operational baselines and existing efficient AI baselines. In practice, Sonny converges in approximately 5.5 days on a single NVIDIA A40 GPU.

Future work can extend Sonny in several directions. First, multi-path inference from randomized intervals can be leveraged to estimate forecast uncertainty in a principled way. Second, we will integrate land-surface and oceanic variables into the forecasting pipeline and evaluate whether this coupled setting improves robustness and forecast accuracy. Third, evaluating Sonny at higher spatial resolutions and larger model scales is a promising direction, particularly given the favorable efficiency-accuracy trade-off observed in this work.

\bibliographystyle{unsrtnat}
\bibliography{cas-refs}

@article{Pathak2022FourCastNet,
  author  = {Pathak, Jaideep and Subramanian, Shashank and Harrington, Peter and others},
  title   = {FourCastNet: A Global Data-driven High-Resolution Weather Model using Adaptive Fourier Neural Operators},
  journal = {arXiv preprint arXiv:2202.11214},
  year    = {2022}
}

@article{Bi2023Pangu,
  author  = {Bi, Kaifeng and Xie, Lingxi and Zhang, Hengheng and Chen, Xin and Gu, Xiaotao and Tian, Qi},
  title   = {Accurate medium-range global weather forecasting with 3D neural networks},
  journal = {Nature},
  volume  = {619},
  number  = {7970},
  pages   = {533--538},
  year    = {2023}
}

@article{Lam2023GraphCast,
  author  = {Lam, Remi and Sanchez-Gonzalez, Alvaro and Willson, Matthew and others},
  title   = {Learning skillful medium-range global weather forecasting},
  journal = {Science},
  volume  = {382},
  number  = {6677},
  pages   = {1416--1421},
  year    = {2023}
}

@misc{IPCC2021,
  author       = {{IPCC}},
  title        = {Climate Change 2021: The Physical Science Basis. Contribution of Working Group I to the Sixth Assessment Report of the Intergovernmental Panel on Climate Change},
  year         = {2021},
  publisher    = {Cambridge University Press},
  address      = {Cambridge, United Kingdom and New York, NY, USA},
  doi          = {10.1017/9781009157896}
}

@article{dunstan2025fastnet,
  title={FastNet: Improving the physical consistency of machine-learning weather prediction models through loss function design},
  author={Dunstan, Tom and Strickson, Oliver and Bennett, Thusal and Bowyer, Jack and Burnand, Matthew and Chappell, James and Coca-Castro, Alejandro and Dale, Kirstine Ida and Daub, Eric G and Eftekhari, Noushin and others},
  journal={arXiv preprint arXiv:2509.17601},
  year={2025}

}

@article{khadir2025democracy,
  title={Democracy of ai numerical weather models: An example of global forecasting with fourcastnetv2 made by a university research lab using gpu},
  author={Khadir, Iman and Stevenson, Shane and Li, Henry and Krick, Kyle and Burrows, Abram and Hall, David and Posey, Stan and Shen, Samuel SP},
  journal={arXiv preprint arXiv:2504.17028},
  year={2025}
}

@article{han2025step,
  title={Step by Step Network},
  author={Han, Dongchen and Ye, Tianzhu and Xia, Zhuofan and Chen, Kaiyi and Wang, Yulin and Chen, Hanting and Huang, Gao},
  journal={arXiv preprint arXiv:2511.14329},
  year={2025}
}

@article{bonev2023sfno,
  title={Spherical Fourier Neural Operators: Learning Stable Dynamics on the Sphere},
  author={Bonev, Boris and Kurth, Thorsten and Hundt, Christian and Pathak, Jaideep and Baust, Maximilian and Kashinath, Karthik and Anandkumar, Anima},
  journal={arXiv preprint arXiv:2306.03838},
  year={2023}
}

@article{couairon2024archesweather,
  title={Archesweather: An efficient ai weather forecasting model at 1.5 $\\{\\backslashdeg\\}$ resolution},
  author={Couairon, Guillaume and Lessig, Christian and Charantonis, Anastase and Monteleoni, Claire},
  journal={arXiv preprint arXiv:2405.14527},
  year={2024}
}

@article{rasp2024weatherbench2,
  title={WeatherBench 2: A Benchmark for the Next Generation of Data-Driven Global Weather Forecasting Models},
  author={Rasp, w and Thuerey, Nils and others},
  journal={arXiv preprint arXiv:2308.15560},
  year={2024}
}

@article{nguyen2024scaling,
  title={Scaling transformer neural networks for skillful and reliable medium-range weather forecasting},
  author={Nguyen, Tung and Shah, Rohan and Bansal, Hritik and Arcomano, Troy and Maulik, Romit and Kotamarthi, Rao and Foster, Ian and Madireddy, Sandeep and Grover, Aditya},
  journal={Advances in Neural Information Processing Systems},
  volume={37},
  pages={68740--68771},
  year={2024}
}

@article{polyak1992acceleration,
  title={Acceleration of stochastic approximation by averaging},
  author={Polyak, Boris T and Juditsky, Anatoli B},
  journal={SIAM Journal on Control and Optimization},
  volume={30},
  number={4},
  pages={838--855},
  year={1992}
}

@article{tarvainen2017meanteacher,
  title={Mean teachers are better role models: Weight-averaged consistency targets improve semi-supervised deep learning results},
  author={Tarvainen, Antti and Valpola, Harri},
  journal={Advances in Neural Information Processing Systems},
  volume={30},
  year={2017}
}

@article{grill2020byol,
  title={Bootstrap your own latent: A new approach to self-supervised learning},
  author={Grill, Jean-Bastien and Strub, Florian and Altch{\'e}, Florent and Tallec, Corentin and Richemond, Pierre H. and Buchatskaya, Elena and Doersch, Carl and Pires, Bernardo {\'A}vila and Guo, Zhaohan Daniel and Azar, Mohammad Gheshlaghi and others},
  journal={Advances in Neural Information Processing Systems},
  volume={33},
  pages={21271--21284},
  year={2020}
}

@article{ahn2023searching,
  title={Searching similar weather maps using convolutional autoencoder and satellite images},
  author={Ahn, Heewoong and Lee, Sunhwa and Ko, Hanseok and Kim, Meejoung and Han, Sung Won and Seok, Junhee},
  journal={ICT Express},
  volume={9},
  number={1},
  pages={69--75},
  year={2023},
  publisher={Elsevier}
}

@article{bodnar2025foundation,
  title={A foundation model for the Earth system},
  author={Bodnar, Cristian and Bruinsma, Wessel P and Lucic, Ana and Stanley, Megan and Allen, Anna and Brandstetter, Johannes and Garvan, Patrick and Riechert, Maik and Weyn, Jonathan A and Dong, Haiyu and others},
  journal={Nature},
  volume={641},
  number={8065},
  pages={1180--1187},
  year={2025},
  publisher={Nature Publishing Group UK London}
}

@article{chen2026emformer,
  title={EMFormer: Efficient Multi-Scale Transformer for Accumulative Context Weather Forecasting},
  author={Chen, Hao and Han, Tao and Zhang, Jie and Guo, Song and Ling, Fenghua and Bai, Lei},
  journal={arXiv preprint arXiv:2602.01194},
  year={2026}
}

@article{cheon2024karina,
  title={Karina: An efficient deep learning model for global weather forecast},
  author={Cheon, Minjong and Choi, Yo-Hwan and Kang, Seon-Yu and Choi, Yumi and Lee, Jeong-Gil and Kang, Daehyun},
  journal={arXiv preprint arXiv:2403.10555},
  year={2024}
}

@article{peebles2023dit,
  title={Scalable Diffusion Models with Transformers},
  author={Peebles, William and Xie, Saining},
  journal={Proceedings of the IEEE/CVF International Conference on Computer Vision},
  pages={4195--4205},
  year={2023}
}

@article{han2024fengwu,
  title={Fengwu-ghr: Learning the kilometer-scale medium-range global weather forecasting},
  author={Han, Tao and Guo, Song and Ling, Fenghua and Chen, Kang and Gong, Junchao and Luo, Jingjia and Gu, Junxia and Dai, Kan and Ouyang, Wanli and Bai, Lei},
  journal={arXiv preprint arXiv:2402.00059},
  year={2024}
}

@article{bonev2025fourcastnet,
  title={Fourcastnet 3: A geometric approach to probabilistic machine-learning weather forecasting at scale},
  author={Bonev, Boris and Kurth, Thorsten and Mahesh, Ankur and Bisson, Mauro and Kossaifi, Jean and Kashinath, Karthik and Anandkumar, Anima and Collins, William D and Pritchard, Michael S and Keller, Alexander},
  journal={arXiv preprint arXiv:2507.12144},
  year={2025}
}

@article{cheon2025modernizing,
  title={Modernizing CNN-based Weather Forecast Model towards Higher Computational Efficiency},
  author={Cheon, Minjong and Goo, Eunhan and Shin, Su-Hyeon and Ahmed, Muhammad and Kim, Hyungjun},
  journal={arXiv preprint arXiv:2507.10893},
  year={2025}
}

\section{Supplementary Forecast Visualization}

For supplementary qualitative analysis, we additionally generated global forecast visualizations over lead times from day 1 through day 14 for nine core meteorological variables. All rollouts start from the same initialization time (00:00 UTC, January 26, 2020) to ensure consistent cross-lead-time comparison. Each panel corresponds to one target lead time, and each row corresponds to one variable. The four columns are organized as follows: initial condition, ERA5 reference at the target lead time, Sonny prediction, and prediction bias (forecast minus reference). In the main supplementary set, we provide representative examples at day 1, day 10, and day 14 to highlight short-range fidelity, medium-range stability, and long-range error structure.

\begin{figure*}[htbp]
\centering
\includegraphics[width=0.85\textwidth]{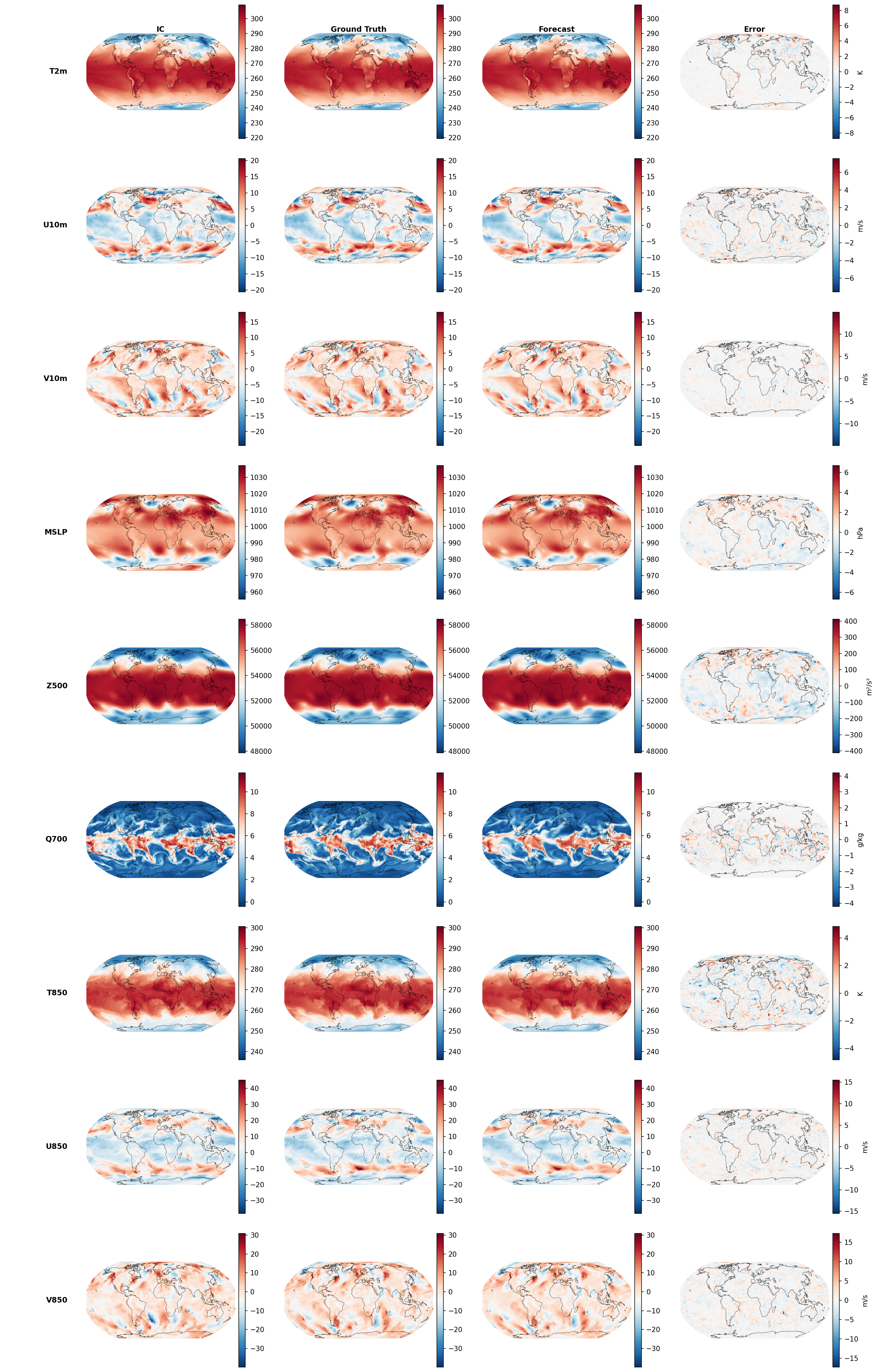}
\caption{Supplementary forecast visualization at day 1.}
\label{fig:supp_day1}
\end{figure*}

\begin{figure*}[htbp]
\centering
\includegraphics[width=0.85\textwidth]{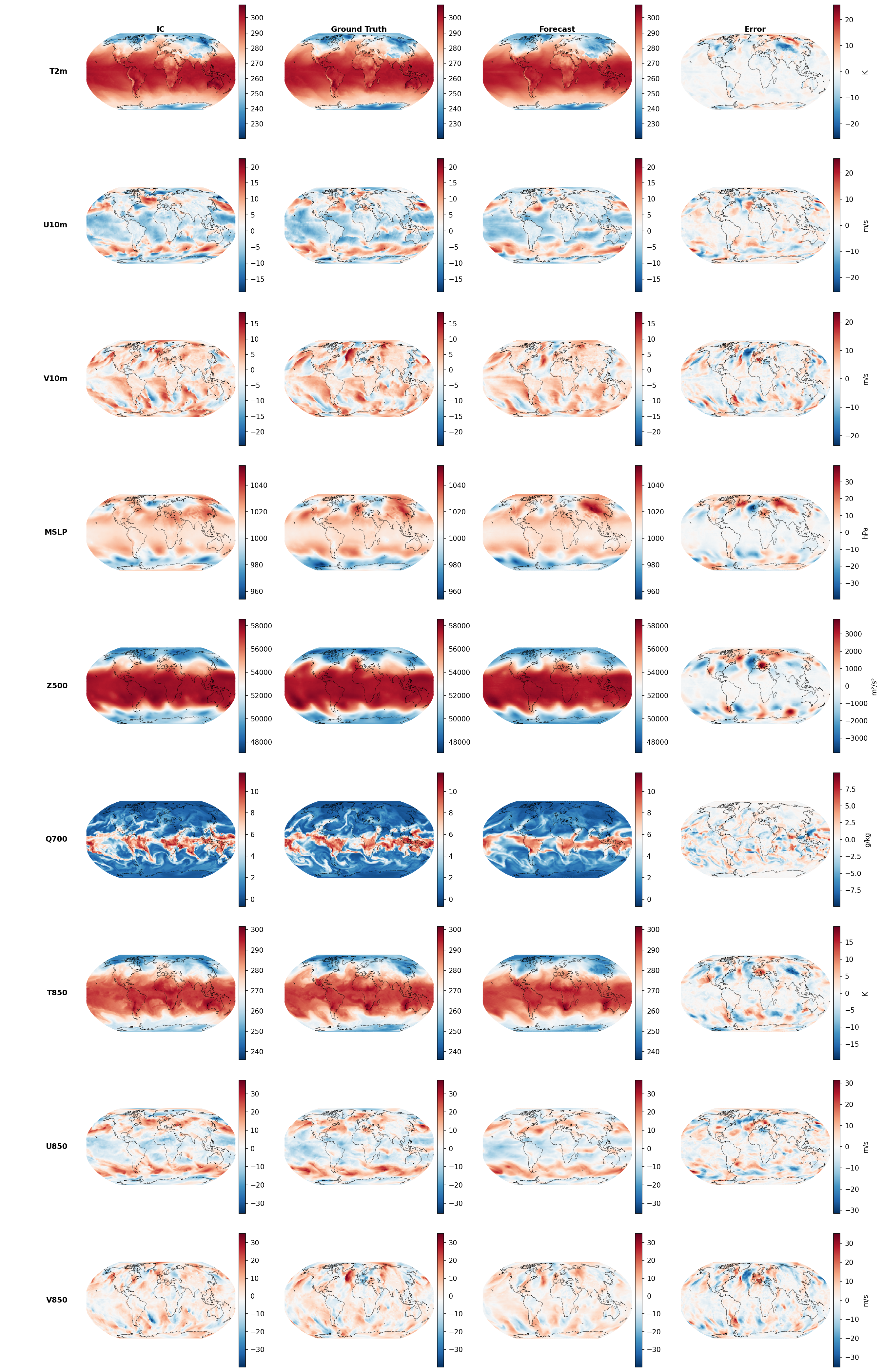}
\caption{Supplementary forecast visualization at day 10.}
\label{fig:supp_day10}
\end{figure*}

\begin{figure*}[htbp]
\centering
\includegraphics[width=0.85\textwidth]{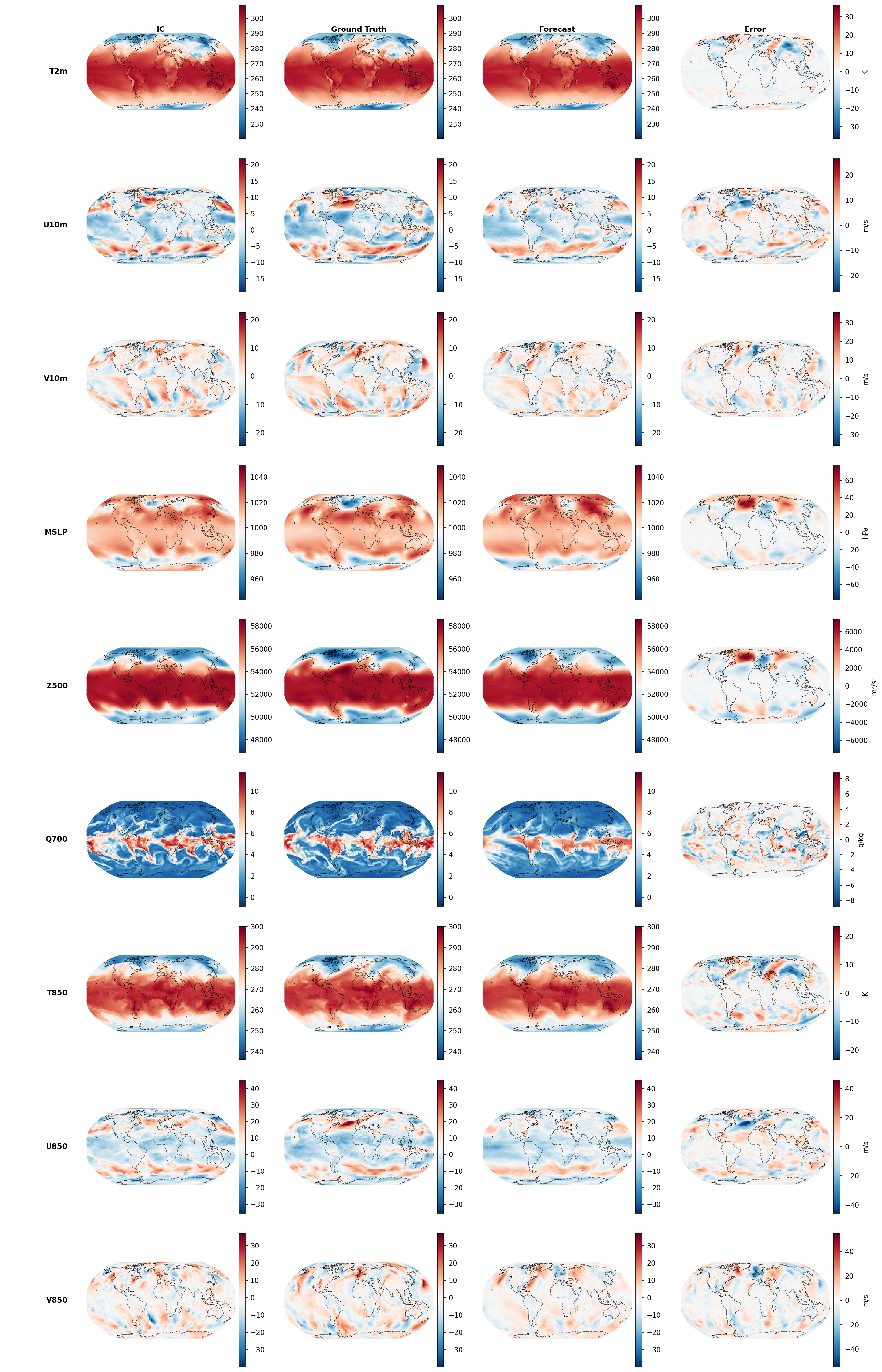}
\caption{Supplementary forecast visualization at day 14.}
\label{fig:supp_day14}
\end{figure*}

\end{document}